\documentclass[letterpaper, 10 pt, conference]{ieeeconf}  

\IEEEoverridecommandlockouts                              
 
\usepackage{graphics} 
\usepackage{epsfig} 
\usepackage{mathptmx} 
\usepackage{times} 
\usepackage{amsmath} 
\usepackage{amssymb}  
\usepackage{arydshln}
\usepackage{booktabs}
\usepackage{pifont}
\usepackage{tabularx}
\usepackage[table]{xcolor}
\usepackage[dvipsnames]{xcolor}
\usepackage{verbatim}
\usepackage{multirow}
\usepackage{makecell}
\usepackage{hyperref}
\usepackage{cleveref}
\usepackage{cite}
\usepackage{pifont}

\usepackage{balance}

\title{\LARGE \bf
Dexora: Open-source VLA for High-DoF Bimanual Dexterity
}
\author{Zongzheng Zhang$^{1, 2*}$, Jingrui Pang$^{1, 2*}$, Zhuo Yang$^{1}$, Kun Li$^{2}$, Minwen Liao$^{1}$, \\Saining Zhang$^{1}$, Guoxuan Chi$^{1}$, Jinbang Guo$^{2}$, Huan-ang Gao$^{1}$, Modi Shi$^{3}$, \\Dongyun Ge$^{1}$, Yao Mu$^{4}$, Jiayuan Gu$^{5}$, Rui Chen$^{1}$, Hao Dong$^{6}$, Huazhe Xu$^{1}$, Li Yi$^{1}$, Yixin Zhu$^{6}$, \\Hang Zhao$^{1}$, Pengwei Wang$^{2}$, Shanghang Zhang$^{2, 6}$, Guocai Yao$^{2}$, Jianyu Chen$^{1}$, Hongyang Li$^{3}$, Hao Zhao$^{1, 2\dagger}$%
\thanks{%
    $^{1}$Tsinghua University.
    $^{2}$Beijing Academy of Artificial Intelligence.
    $^{3}$The University of Hong Kong.
    $^{4}$Shanghai Jiao Tong University.
    $^{5}$ShanghaiTech University.
    $^{6}$Peking University.
    $^{*}$Equal contribution. $^{\dagger}$ Corresponding author%
}%
}

\begin{document}

\maketitle
\thispagestyle{empty}
\pagestyle{empty}

\begin{abstract}

Vision-Language-Action (VLA) models have recently become a central direction in embodied AI, but current systems are restricted to either dual-gripper control or single-arm dexterous hand manipulation. While low-dimensional gripper control can often be handled with simpler methods, high-dimensional dexterous hand control benefits greatly from full end-to-end VLA learning. In this work, we introduce Dexora, the first open-source VLA system that natively targets dual-arm, dual-hand high-DoF manipulation. We design a hybrid teleoperation pipeline that decouples gross arm kinematics (captured with a custom exoskeleton backpack) from fine finger motion (markerless hand tracking via Apple Vision Pro), and that drives both a physical dual-arm dual-hand platform and an identical MuJoCo digital twin. Using that interface, we assemble a large training corpus: an embodiment-matched synthetic corpus (100K simulated trajectories, 6.5M frames) and a real-world dataset of 10K teleoperated episodes (2.92M frames). To mitigate noisy teleoperation demonstrations, we propose a data-quality-aware training recipe: an offline discriminator provides clip-level weights for diffusion-transformer policy training, down-weighting low-quality demonstrations. Empirically, Dexora outperforms competitive VLA baselines on both basic and dexterous benchmarks (e.g., average dexterous success 66.7\% vs. 51.7\%), attains 90\% success on basic tasks, and shows robust out-of-distribution and cross-embodiment generalization. Ablations confirm the importance of real data and the discriminator for dexterity. Demos, data, code, and models can be found at \href{https://dexoravla.github.io}{\textcolor{orange}{\texttt{https://dexoravla.github.io}}}.

\end{abstract}
\section{Introduction}

\begin{figure}
    \centering
    \includegraphics[width=1.0\linewidth]{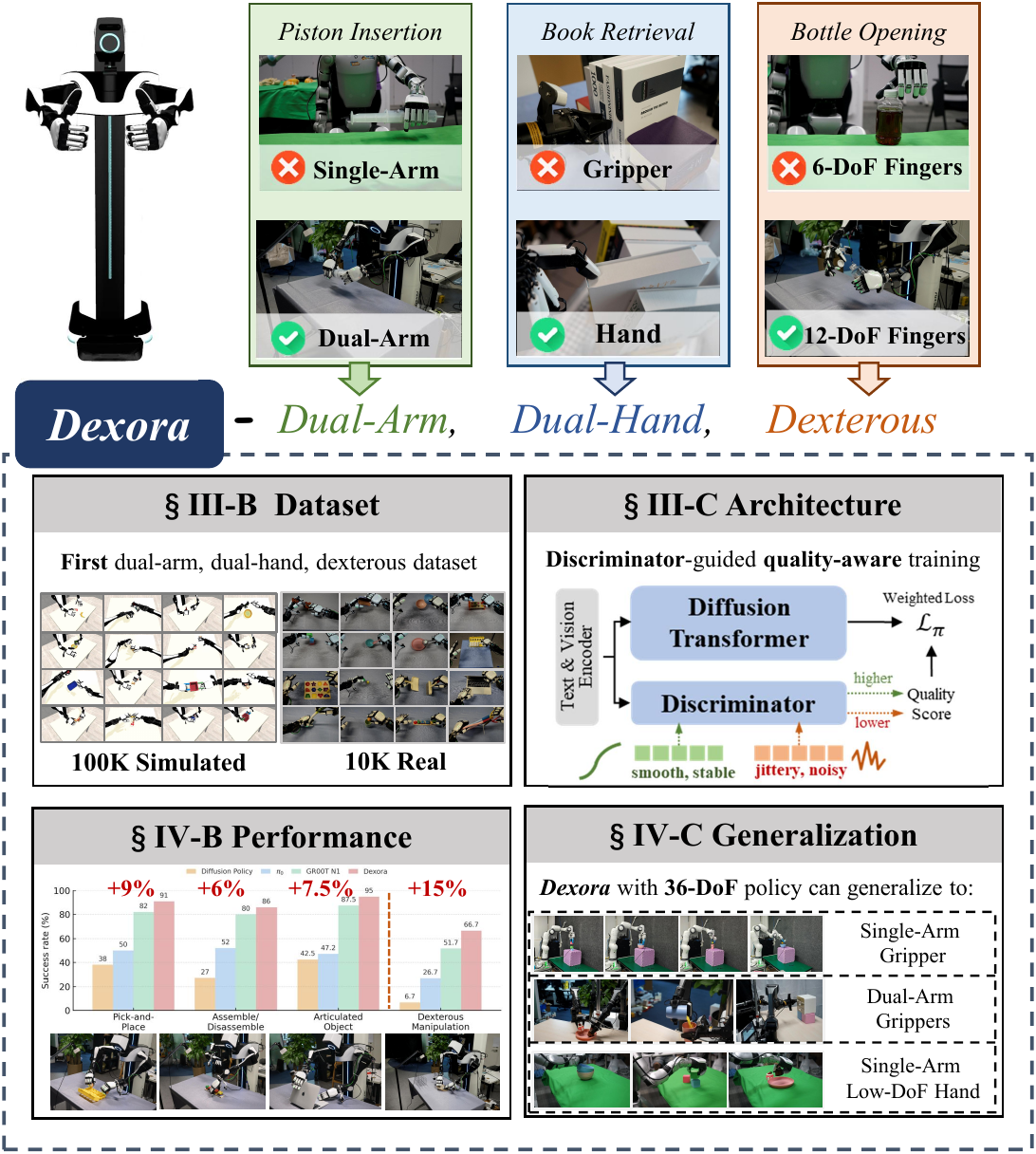}
    \caption{\textbf{\emph{Dexora} overview}. (a) \textbf{Motivation}: Three illustrative contrasts highlight the need for dual-arm, dual-hand dexterous VLA: piston insertion (requires two arms), book retrieval from a packed shelf (hands with fingers succeed where grippers fail), and bottle opening (12-DoF fingers with lateral swing outperform 6-DoF). (b) \textbf{Dataset} (§\ref{sec: method-dataset}): We pretrain on \textbf{100K} simulated bimanual-hand trajectories and post-train on \textbf{10K} real demonstrations, all collected with our dual-arm, dual-hand platform.
   (c) \textbf{Architecture} (§\ref{sec: method-framework}): A trained discriminator scores dataset demonstration quality and guides training, driving the diffusion-transformer policy to prioritize high-quality trajectories while down-weighting low-quality ones.
     (d) \textbf{Performance} (§\ref{sec: real world results}): \emph{Dexora} achieves consistently higher average success rates on both basic (Pick-and-Place, Assemble/Disassemble, Articulated Object) and dexterous benchmarks compared to state-of-the-art VLA models. (e) \textbf{Embodiment generalization} (§\ref{sec: generalization}): The same policy transfers across \textbf{single-arm gripper}, \textbf{dual-arm grippers}, and \textbf{single-arm low-DoF hand} without re-architecting the model.}
     \vspace{-10pt}
    \label{fig:teaser}
\end{figure}

\begin{figure}
    \centering
    \includegraphics[width=1.0\linewidth]{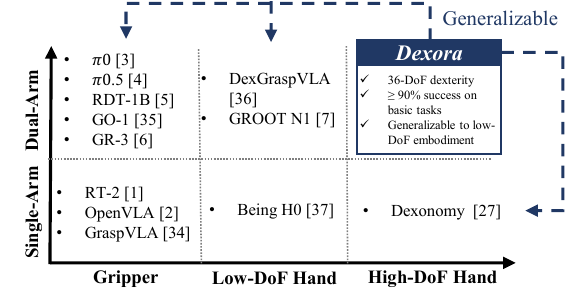}
    \caption{\textbf{Comparison of embodiment coverage.} Prior works cover either single-arm or low-DoF dual-arm settings. \emph{Dexora} is the first system positioned in the dual-arm, high-DoF dexterous region, while also generalizing across simpler embodiments without re-architecture.}
    \label{fig:intro-6areas}
    \vspace{-0.5cm}
\end{figure}

Vision-Language-Action (VLA) models have emerged as a promising paradigm for embodied AI, yet existing systems remain fundamentally constrained: they are either designed for dual-arm, low-DoF grippers or single-arm dexterous hands, but not both~\cite{zitkovich2023rt2, kim2024openvla, black2024pi_0, intelligence2025pi_05, liu2024rdt, gr3, bjorck2025gr00t}. As illustrated in Fig.~\ref{fig:teaser} (top), such limitations prevent prior VLAs from handling tasks that intrinsically demand dual-arm coordination (e.g., piston insertion), or high-DoF dexterous fingers (e.g., bottle opening/complex book retrieval). \emph{Dexora} is the first open-source VLA that addresses this gap by unifying dual-arm, dual-hand, and high-DoF dexterity into a single system (Fig.~\ref{fig:intro-6areas}).

To enable such complex skill acquisition, \emph{Dexora} introduces a hybrid teleoperation pipeline. Gross arm kinematics are captured with a lightweight exoskeleton backpack, while fine-grained finger articulation is driven by markerless hand tracking via Apple Vision Pro. This decoupling makes it feasible to control a physical dual-arm dual-hand platform with 36 DoF, while simultaneously mirroring demonstrations in a MuJoCo-based digital twin, thereby ensuring scalable and embodiment-matched data collection.

Using this interface, we construct a large-scale dataset for dual-arm, dual-hand dexterous manipulation (Fig.~\ref{fig:teaser}, §\ref{sec: method-dataset}). It consists of ~100K simulated trajectories (361 hours, 6.5M frames) and 10K real teleoperated episodes (177.5 hours, 3.2M frames). The design follows the principle of sim-real complementarity: simulated data provide scale and task diversity, while real data provides fine-grained realism essential for high-DoF bimanual dexterity. Together, this dataset establishes a foundation for training VLA models under realistic dexterous settings.

A key challenge of teleoperated data is the presence of noisy or unstable demonstrations (Fig.~\ref{fig:teaser}, §\ref{sec: method-framework}). To address this, \emph{Dexora} employs discriminator-guided quality-aware training: an offline discriminator scores each demonstration, and the policy is trained with weighted diffusion-transformer loss that down-weights low-quality clips. This design effectively stabilizes learning, ensuring that the policy benefits from large-scale data while mitigating the impact of teleoperation artifacts.

We evaluate \emph{Dexora} across both basic manipulation and dexterous benchmarks (Fig.~\ref{fig:teaser}, §\ref{sec: real world results}). Quantitatively, \emph{Dexora} achieves over 90\% success on basic pick-and-place and open articulated objects tasks, while improving dexterous success from 51.7\% (baseline) to 66.7\% (+15\%). Qualitatively, the system demonstrates torsional manipulation and complex dual-arm coordination. These results highlight the critical role of both real-world data and quality-aware training in attaining high-DoF dexterity.

Finally, \emph{Dexora} exhibits strong generalization beyond its native embodiment (Fig.~\ref{fig:teaser}, §\ref{sec: generalization}). Despite being trained on a 36-DoF dual-arm dual-hand platform, the learned policy successfully transfers to single-arm gripper, dual-arm grippers, and single-arm low-DoF hand. This suggests that VLA policies trained under rich dexterous settings can serve as universal controllers, generalizing across embodiments.
Fig.~\ref{fig:intro-6areas} situates this result in the broader landscape: prior VLAs mainly focus on single-/dual-arm grippers or low-DoF hand. \emph{Dexora} is positioned in the dual-arm, high-DoF hands quadrant while remaining downward-compatible to the other regions of the grid. This suggests a practical route to universal controllers: train in the dexterous, high-DoF setting and deploy by projecting to simpler robots.

\section{Related Work}

\subsection{Teleoperation System}

Teleoperation enables us to acquire large-scale robot demonstrations by translating human motions into robot-executable control signals. Existing platforms can be categorized into five classes: (i) leader–follower systems with kinesthetic teaching rigs~\cite{zhao2023learningfinegrained,aloha22024enhanced}; (ii) VR/MR headset–based pose tracking (e.g., Vision Pro pipelines)~\cite{iyer2024openteach,ding2024bunnyvisionpro}; (iii) vision-only retargeting~\cite{qin2023anyteleop,li2018teachnet}; (iv) exoskeleton interfaces for joint-level arm and finger tracking~\cite{fang2023airexo,xu2025dexumi}; and (v) joystick/button controllers~\cite{imdieke2025sparkremote,wu2023gello}. We adopt a hybrid teleoperation setup: exoskeletons provide precise arm-level kinematics, while the Vision Pro offers convenient, high-resolution capture of fine-finger motions. This combination produces high-DoF, dual-arm and dual-hand demonstrations that are both accurate and operator-friendly, and are natively compatible with Vision-Language-Action (VLA) model training~\cite{li2025demonstrationmodality,pan2024vladiffswitch}.

\subsection{Dexterous Manipulation}

Dexterous manipulation includes grasping, in-hand reconfiguration, tool use, and coordinated bi-manual skills ~\cite{zhang2024dexgraspnet, ye2025dex1b}. Prior research generally falls into two categories: \emph{grasp synthesis} and \emph{policy learning}. \textbf{On the synthesis side}, the field has undergone a paradigm shift from analytical sampling to generative modeling. Diffusion~\cite{ye2024g,2025dexgraspanything}, normalizing flows~\cite{xu2023unidexgrasp}, and latent generative models such as VAE~\cite{li2024semgrasp,liu2024realdex}, complemented by optimization-based pipelines~\cite{chen2025dexonomy}, now enable scalable production of physically consistent grasps across diverse hands and objects. \textbf{On the policy side}, reinforcement learning~\cite{zhang2025robustdexgrasp} and imitation learning~\cite{li2025maniptrans} have driven progress toward closed-loop robustness and sim-to-real transfer in high-DoF hands. Emerging \emph{data engines} leverage automated imitation~\cite{jiang2024dexmimicgen} and egocentric supervision~\cite{yang2025egovla} to expand coverage, accelerating policy learning at unprecedented scale. Despite this rapid progress, most pipelines remain hand-centric, reward-sensitive, and limited in multi-arm coordination. In contrast, we pursue a vision-language-action (VLA) model that operates in \textbf{dual-arm, dual-hand} high-dimensional action space.

\subsection{Vision-Language-Action (VLA) Model}

Vision-Language-Action (VLA) models have recently emerged as a promising paradigm yet most existing systems remain confined to low-DoF or single-arm embodiments~\cite{zhang2025tavla,zhang2025robochemist}. Representative efforts such as RT-2~\cite{zitkovich2023rt2}, OpenVLA~\cite{kim2024openvla}, and GraspVLA~\cite{2025graspvla} output manipulation policies for single-arm grippers. More recent generalist policies extend to bimanual settings—e.g., $\pi_0$~\cite{black2024pi_0}, $\pi_{0.5}$~\cite{intelligence2025pi_05}, RDT~\cite{liu2024rdt}, GO-1~\cite{bu2025agibot}, GR-3~\cite{gr3}, GR00T~\cite{bjorck2025gr00t}, and DexGraspVLA~\cite{2025dexgraspvla}—but these typically simplify embodiment to parallel-jaw grippers, limiting dexterity. In parallel, large-scale data engines such as Being-H0~\cite{luo2025beingH0} and DreamGen~\cite{jang2025dreamgen} have enriched supervision, but they still fall short of enabling \textbf{high-DoF} dual-hand control.

Our work introduces a \textbf{dual-arm, dual-hand high-DoF VLA} that learns to output synchronized arm–hand trajectories end-to-end. The formulation admits natural downshifting to lower-DoF embodiments via finetuning, offering a unified pathway toward cross-embodiment generalization.

\begin{figure*}
    \centering
    \includegraphics[width=0.9\linewidth]{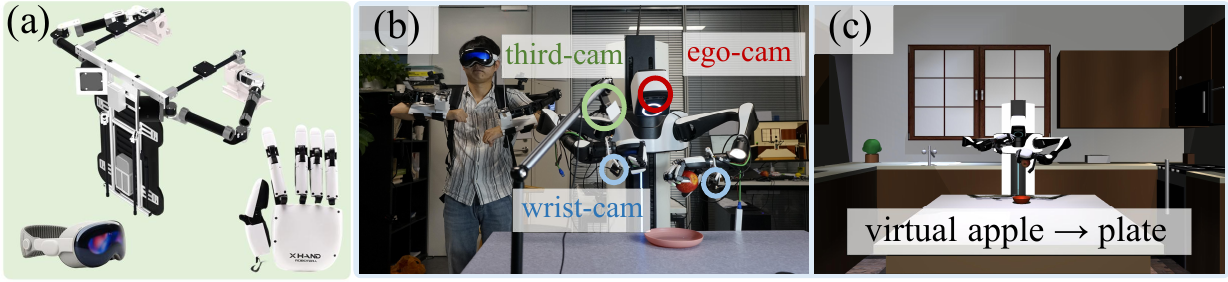}
    \caption{\textbf{Hardware and teleoperation system}. (a) Hybrid teleoperation interface and 12-DoF XHAND. (b)-(c) The operator teleoperates the physical robot and its MujoCo digital twin, so \emph{apple$\rightarrow$plate} demonstrations are collected in real and simulation under the same interface, thereby reducing the sim-to-real gap.}
    \vspace{-10pt}
    \label{fig:hard-tele}
\end{figure*}

\section{Dexora}

In this section, we first introduce the hardware setup and teleoperation system (Sec.~\ref{sec: method-hardware system}), followed by the construction of our dataset, assembling an embodiment-aligned corpus of large-scale synthetic and real-world demonstrations (Sec.~\ref{sec: method-dataset}). We then present the VLA framework with a learned data-quality discriminator that scores demonstrations and weights training (Sec.~\ref{sec: method-framework}). Finally, we specify the three-stage data-quality-aware training recipe (Sec.~\ref{sec: method-training recipe}).

\subsection{Dual-Arm Dual-hand System}
\label{sec: method-hardware system}

As shown in Fig.~\ref{fig:hard-tele} (a), \emph{Dexora} integrates two 6-DoF AIRBOT arms with a pair of XHAND dexterous hands, each offering 12 fully actuated joints. All finger joints are independently driven, and the thumb and index additionally support lateral ab/adduction, enabling human-like in-hand reorientation and torsional manipulation (e.g., cap twisting).

To achieve scalable teleoperation, we decouple gross arm motion from fine finger control. A custom dual-arm exoskeleton backpack captures the operator’s shoulder–elbow–wrist angles and maps them directly to robot joint space. This design yields drift-free, low-latency trajectories while avoiding the inverse-kinematics jitter and singularities that often degrade vision-only retargeting pipelines. Apple Vision Pro provides markerless 3D finger skeletons that we retarget to XHAND with a short calibration phase while enforcing joint limits and safety constraints. This hybrid interface combines the precision of joint-space control for the arms and the convenience of lightweight, glove-free finger input, making long data-collection sessions practical (Fig.~\ref{fig:hard-tele} (a)).

Our interface drives both the physical robot and a MuJoCo digital twin of the same embodiment. All sensing streams share a time-aligned I/O system: four RGB views and full 36-DoF joint states are logged at 20 Hz. The twin mirrors the real robot’s kinematics and controllers, and the same teleop drivers run in real and sim, yielding low latency and high fidelity; operators can switch seamlessly between hardware and simulation to collect demonstrations (Fig.~\ref{fig:hard-tele} (b)-(c)).


\subsection{Dataset Construction}
\label{sec: method-dataset}

\begin{figure}
    \centering
    \includegraphics[width=1.0\linewidth]{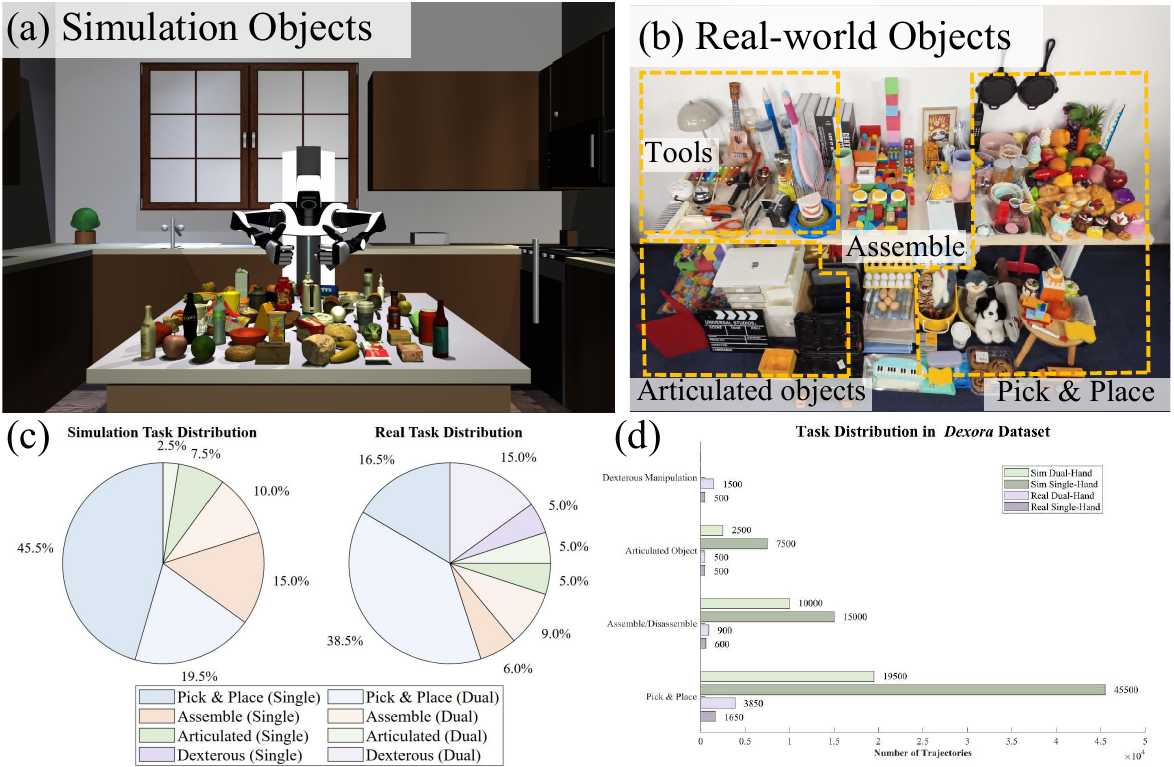}
    \caption{\textbf{Dataset demonstration}. (a) Simulation objects subset: our simulator includes 297 objects across 30 categories. (b) Real-world objects (347 objects, 17 categories), covering both basic and dexterous use cases. (c) Per-family task distribution in simulation vs. real. The simulation data only includes basic tasks, while the real-world set shifts weight toward dexterity (20\%). (d) Trajectory counts per family and embodiment (sim/real; single-/dual-hand).}
    \vspace{-20pt}
    \label{fig:dataset demonstration}
\end{figure}

\textbf{Synthetic Data.}
We generate a large, embodiment-matched simulation corpus in MuJoCo. Using Qwen2.5-VL~\cite{bai2025qwen2}, we mine Objaverse~\cite{deitke2023objaverse} to select manipulable objects and automatically assign physical parameters (Fig.~\ref{fig:dataset demonstration} (a)). On top of this, we build a set of 200 tasks covering three basic families in Fig.~\ref{fig:dataset demonstration} (c). For each task, we collect 3–5 teleoperated seed demonstrations and follow the DexMimicGen~\cite{jiang2024dexmimicgen} recipe to synthesize trajectories: we randomize initial states and retarget the seed actions to new scenes, yielding 500 trajectories per task. Scene layouts and success criteria are auto-generated by Qwen. All simulated episodes are logged with the same observation–action protocol as in the real system, which keeps the interface consistent and reduces the sim-to-real gap. In total, the synthetic set contains about 6.5M frames, 361h video. 

\textbf{Real World Data.}
We collect real-world data on the same embodiment used in the simulation. Beyond common objects and basic tasks, we add dexterous tool-use scenarios that are difficult to stage in simulation (Fig.~\ref{fig:dataset demonstration} (b)) and the dexterous scenes in Fig.~\ref{fig:dataset demonstration} (c)–(d). In total, we curate 200 tasks and acquire 50 teleoperated demonstrations per task via the hybrid teleoperation interface, yielding 10K episodes. The dataset amounts to 40.5 hours and 2.92M frames. All recordings are converted to the LIBERO-2.1 standard and open source. We use this to fine-tune the VLA to specialize basic competence into dexterous, bimanual skills.

\subsection{Framework}
\label{sec: method-framework}


\begin{figure*}
    \centering
    \includegraphics[width=0.92\linewidth]{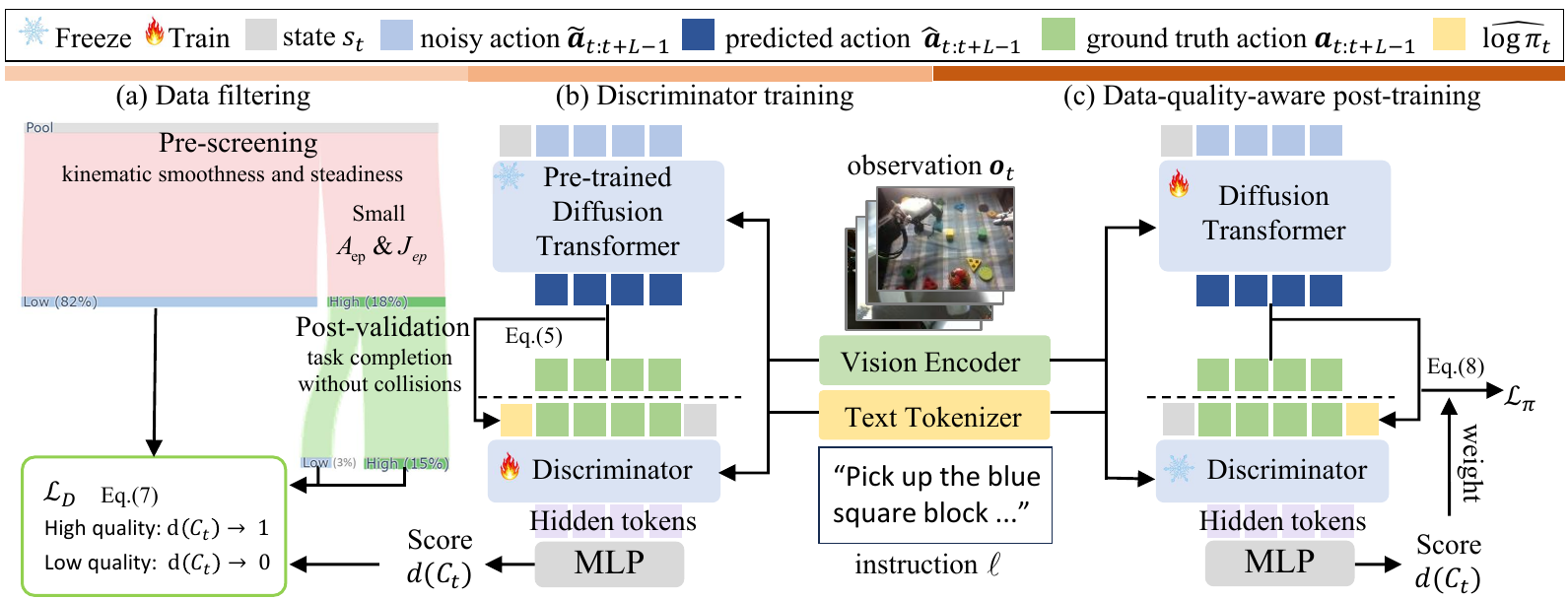}
    \caption{\textbf{\emph{Dexora} framework.} (a) \textbf{Data filtering}: From the real-world dataset we pre-screen demonstrations by kinematic smoothness (low acceleration and jerk), then replay them for post-validation and keep the clips that complete the task without collisions, forming a high-quality subset. (b) \textbf{Discriminator training}: With the pretrained diffusion–transformer policy frozen, we compute a log-$\pi$ proxy for each clip and train a discriminator that, conditioned on observations and language, outputs a quality score $d(C_t)\in(0,1]$. (c) \textbf{Data-quality-aware post-training}: During post-training, the score $d(C_t)$ is converted to weights $w_i$ and used in the diffusion loss $\mathcal{L}_\pi$. At inference time, only the policy is used. }
    \vspace{-10pt}
    \label{fig:method}
\end{figure*}


\textbf{Data Quality Criteria.}
Real-world teleoperation demonstrations exhibit substantial variability due to operator skill, sensing noise, inherent limitations (such as occlusion during hand keypoint tracking), and latency.
Training on such heterogeneous data without constraints often degrades policy learning.
We therefore establish \textbf{episode-level} quality criteria with two pillars: (i) kinematic smoothness and steadiness, proxied by low acceleration \(A_{\text{ep}}\) and jerk \(J_{\text{ep}}\)—for pre-screening; (ii) replay success as the decisive indicator of data reliability (task completion without collisions)—for post-validation. This two-stage design yields a clean positive set for training the discriminator (Fig.~\ref{fig:method} (a)).

Let an episode be denoted by 
\(\tau = \{s_t\}_{t=1}^{T}\), 
where \(s_t \in \mathbb{R}^{D}\) is the proprioceptive state vector (\(D=36\)).
The sampling interval is \(\Delta t\). Because state dimensions have heterogeneous numeric ranges, we first apply per-dimension min--max normalization. We compute velocity, acceleration, and jerk using centered finite differences ($t=4,\dots,T-3$):
\begin{equation}
\begin{aligned}
v_t &= \frac{s_{t+1}-s_{t-1}}{2\Delta t}, \quad 
a_t &= \frac{v_{t+1}-v_{t-1}}{2\Delta t}, \quad 
j_t &= \frac{a_{t+1}-a_{t-1}}{2\Delta t}, \quad
\\
\end{aligned}
\end{equation}
For an episode $\tau$, acceleration and jerk are defined via the root mean square (RMS) across both time and dimensions:
\begin{align}
A_{\text{ep}}(\tau) &= 
\sqrt{\frac{1}{(T-6)D}\sum_{t=4}^{T-3}\sum_{k=1}^{D} a_{t,k}^{2}}, \\
J_{\text{ep}}(\tau) &= 
\sqrt{\frac{1}{(T-6)D}\sum_{t=4}^{T-3}\sum_{k=1}^{D} j_{t,k}^{2}} .
\end{align}
Lower values of $A_{\text{ep}}$ and $J_{\text{ep}}$ indicate smoother, steadier demonstrations.
We rank episodes by $A_{\text{ep}}$ and by $J_{\text{ep}}$ separately, keep the lowest $20\%$ in each list, and take their intersection:
$
\mathcal{S}_{\text{pre}} \;=\; \Big\{\tau:\ \tau\in\text{Low-20\%}(A_{\text{ep}}) \;\wedge\; \tau\in\text{Low-20\%}(J_{\text{ep}})\Big\},
$
which retains about $18\%$ of episodes in our data. From $\mathcal{S}_{\text{pre}}$, we designate positives by open-loop replay success---task completion without collisions:
$
\mathcal{S}_{\text{high}} \;=\; \big\{\tau:\tau\in\mathcal{S}_{\text{pre}}\wedge\text{Success}(\tau)=1\ \wedge\ \text{CollisionFree}(\tau)=1\big\},
$ yielding roughly $15\%$ high-quality demonstrations.
Note that we score quality at the episode not chunk-level: stationary chunks can trivially exhibit low acceleration/jerk yet be uninformative. Episode-level aggregation, paired with a movement-coverage guard, suppresses such false positives and better captures overall stability and task competence.

\textbf{Discriminator Model.}
After selecting the top-quality subset, we use an offline discriminator to score every real episode. For each episode, we uniformly sample $K$ sub-clips $\{C_k\}_{k=1}^K$, and construct a tokenized input per clip: \(
\xi_t=\big(s_t,\ \mathbf{o}_{t},\ \ell,\ \mathbf{a}_{t:t+L-1},\ \widehat{\log\pi}_t\big)
\),
where \(\mathbf{o}_{t}\) are multi-view RGB observations, \(\ell\) is the language instruction, \(\mathbf{a}_{t:t+L-1}\) is an action chunk of length \(L\), and \(\widehat{\log\pi}_t\) is a log-\(\pi\) chunk score (policy-compatibility proxy) computed from the pretrained diffusion policy over that clip.

Given a pretrained diffusion-transformer policy \(\pi_{\theta}\), we define a surrogate for \(\log\pi(\mathbf{a}_{t:t+L-1}\mid \ell,\mathbf{o}_{t})\) via the negative denoising residual energy:
\begin{equation}
E_t = \frac{1}{|\mathcal{S}|\,L}\sum_{s\in\mathcal{S}}\sum_{\tau=t}^{t+L-1}
\left\|\varepsilon_{\theta}\!\left(\mathbf{o}_{\tau},\,\ell,\,\mathbf{a}_{\tau:\tau+L-1},\,s_\tau\right)-\varepsilon\right\|_2^2,
\end{equation}
\begin{equation}
\widehat{\log\pi}_t = -\,\mathrm{zscore}\!\left(E_t\right)=-\frac{E_t - \text{Mean}(E)}{\sqrt{\text{Var}(E)+\varepsilon}},
\end{equation}
where \(\mathcal{S}\) is a small set of diffusion steps. Intuitively, larger \(\widehat{\log\pi}_t\) indicates that the policy explains the chunk better.

Each clip is projected into a token sequence:
\(
[\  s_t;\  \mathbf{a}_{t:t+L-1};\  \widehat{\log\pi}_t\ ],
\)
equipped with learned positional embeddings.
Language and image tokens are concatenated as a condition stream.
A shallow stack of Transformer blocks produces hidden tokens, which are globally averaged and passed through a small MLP head with sigmoid to output a clip score \(d(C_k)\in(0,1]\) (Fig.~\ref{fig:method} (b)).

\begin{figure*}
    \centering
    \includegraphics[width=1.0\linewidth]{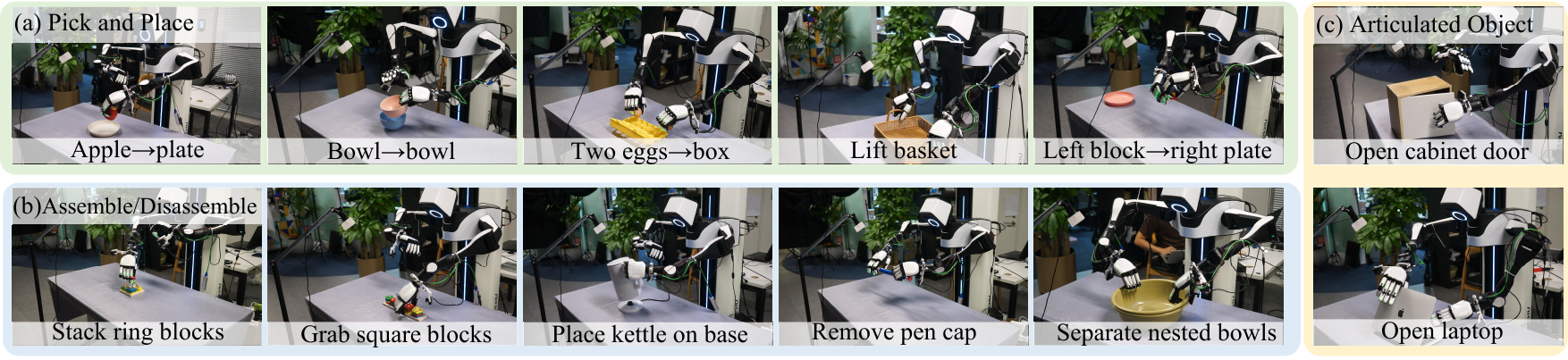}
    \caption{\textbf{Basic tasks suite.} (a) Pick and Place (5 tasks). (b) Assemble/Disassemble (5 tasks). (c) Articulated Objects (2 tasks).}
    \label{fig:basic tasks}
    \vspace{-10pt}
\end{figure*}

\textbf{Diffusion Transformer.}
We employ a decoder-only Transformer as the diffusion model for the policy. Its architecture resembles the discriminator, but the input consists of the current observation $\mathbf{o}_{t}$, and the instruction $\ell$, forming a vision–language conditioned policy:
\begin{equation}
    \pi_{\theta}(s_t,\ \mathbf{o}_{t},\ \ell) = \widehat{\mathbf{a}}_{t:t+L-1}.
\end{equation}
The current joint angle state information state $s_t$, and noisy actions $\widetilde{\mathbf{a}}_{t:t+L-1}$ are projected into the latent space and concatenated with the diffusion timestep \(t\) to form the input tokens for the transformer. Natural language and multi-view image inputs are encoded into conditional tokens via the T5~\cite{T5} and SigLip~\cite{siglip} encoders, respectively, and alternately injected into the transformer blocks. The model predicts the action noise \(\widehat{\theta}\), thereby yielding the predicted action sequence \(\widehat{\mathbf{a}}_{t:t+L-1}\) (Fig.~\ref{fig:method} (c)). We use the standard DDPM for sampling during training and employ DPMSolver++ for acceleration during action generation.

\subsection{Data-quality-aware Training Recipe}
\label{sec: method-training recipe}
We first \textbf{pretrain} the diffusion-transformer policy $\pi_{\theta}$ on simulation data to endow the VLA with basic competence (pick \& place, assemble, etc.).
This policy is then used to compute the \emph{log-$\pi$ proxy} for training the discriminator model.

Let the positive set be the replay-validated high-quality subset
$\mathcal{S}_{\mathrm{high}}$ (about $15\%$) and the unlabeled pool be
$\mathcal{U}=\mathcal{D}_{\mathrm{real}}\setminus \mathcal{S}_{\mathrm{high}}$.
We optimize a positive–unlabeled objective:
\begin{equation}
\mathcal{L}_{D}
= \eta\,\underbrace{\mathbb{E}_{\tau\in \mathcal{S}_{\mathrm{high}}}\!\big[-\log d(\tau)\big]}_{\text{positive BCE}\;\to\;1}
\;+\;
\underbrace{\mathbb{E}_{\tau\in \mathcal{U}}\!\big[-\log (1-d(\tau))\big]}_{\text{unlabeled as negative}\;\to\;0},
\end{equation}
where $\eta=0.5$. We apply clip scores to
$d\in[0.1,0.9]$ for stability (Fig.~\ref{fig:method} (b)).
Following the DWBC mapping from~\cite{xu2022discriminator}, we convert calibrated scores to weights $w_i$.

Finally, we \textbf{post-train} $\pi_{\theta}$ on the real dataset to upgrade this base competence into dexterous skills, using the precomputed weights. For diffusion training,
\begin{equation}
\mathcal{L}_{\pi}
=\sum_{i=1}^{L} w_i\;\big\|\varepsilon_{\theta}(\cdot)-\varepsilon\big\|_2^2,
\end{equation}
with a short weight warm-up (Fig.~\ref{fig:method} (c)).

\section{EXPERIMENT}

We evaluate \emph{Dexora} across three axes: (1) \textbf{Performance}: higher success on basic and dexterous tasks, especially on bimanual skills (Sec.~\ref{sec: real world results}). (2) \textbf{Generalization}: Robust to OOD shifts and transfers across embodiments (Sec.~\ref{sec: generalization}). (3) \textbf{Ablations}:  contributions of training data composition and the learned data-quality discriminator (Sec.~\ref{sec: ablation study}).

\subsection{Experimental Setup and Baselines}

\textbf{Setup.} Our policy model has 28 layers, a hidden size of 1024, and 16 attention heads. The discriminator is smaller, with 12 layers, a hidden size of 512, and 8 attention heads, for 30M parameters.
We pretrain the policy model for 100K gradient steps and train the discriminator model for 10K steps, using distributed data parallelism across 8 × NVIDIA A100 GPUs with a total batch size of 64. Both models are optimized using AdamW.

\textbf{Baselines.} We compare against three representative baselines: \textbf{Diffusion Policy (DP)}~\cite{chi2023diffusionpolicy}—a conditional denoising policy for visuomotor imitation; \textbf{$\pi_0$}~\cite{black2024pi_0}—a VLA with a flow-matching action generator; and \textbf{GR00T N1}~\cite{bjorck2025gr00t}—an open VLA (VLM + DiT) designed for humanoid control. 

\textbf{Action-space Adaptation.} DP natively regresses continuous actions, so we train it directly on our 36-D vector commands. For $\pi_0$, we append a 2-layer MLP projector that maps each model’s native action output to our 36-D joint command. The projector is factorized by physical groups (L/R arm, L/R hand), and learns the expansion from lower-DoF end-effector outputs to our 12-DoF hands via learned synergies. 

\textbf{Protocol.} All other settings are identical across methods: control frequency, action chunk length \(L=32\), camera intrinsics/extrinsics, and the number of views. For each task, we collect 100 demonstrations to train/fine-tune the baselines for 50K steps. Fine-tuning runs on 4 × NVIDIA L20 GPUs with LoRA; inference is performed on a single RTX 4090. We report the success rate over 20 rollouts per task.

\subsection{Evaluation Results in Real World}
\label{sec: real world results}
\textbf{Basic Tasks Evaluation.} We group basic tasks into three types—\textbf{Pick-and-Place} (5 tasks), \textbf{Assemble/Disassemble} (5 tasks), and \textbf{Articulated Objects} (2 tasks). Each type mixes single-hand and bimanual problems. Representative bimanual examples include placing a distant block into a tray via a two-hand handover with temporal ordering, and separating two stacked bowls that require simultaneous two-hand prying (Fig.~\ref{fig:basic tasks}). \emph{Dexora} is evaluated zero-shot. Results in Tab.~\ref{tab:basic_tasks} show that \emph{Dexora} attains the highest overall success, reaching $\geq\!90\%$ on 7/12 tasks and consistently leading the bimanual tasks. GR00T N1~\cite{bjorck2025gr00t} is competitive on simpler, mostly single-hand tasks. $\pi_0$~\cite{black2024pi_0} degrades most after mapping a gripper-centric action space to high-DoF hands, confirming that the low$\rightarrow$ high DoF mapping is ill-posed without embodiment-matched data. Benefiting from many dual-arm episodes in training, \emph{Dexora} shows clear gains on bimanual coordination while maintaining strong performance. Overall, these trends support our design choice: embodiment-matched, high-DoF data are essential for performance.

\begin{table*}[t]
  \centering
  \small
   \caption{\textbf{Basic tasks evaluation}. Results are success rates (\%) over 20 trials. \textcolor{black}{\colorbox{gray!20}{Gray}} columns indicate bimanual tasks.}
  \setlength{\tabcolsep}{4pt}
  \renewcommand{\arraystretch}{0.98}
  \resizebox{\linewidth}{!}{
  \begin{tabular}{l|cc>{\columncolor{gray!20}}c>{\columncolor{gray!20}}c>{\columncolor{gray!20}}c|ccc>{\columncolor{gray!20}}c>{\columncolor{gray!20}}c|c>{\columncolor{gray!20}}c|c}
    \toprule
    \multirow{2}{*}{\textbf{Method}} &
    \multicolumn{5}{c|}{\textbf{Pick and Place}} &
    \multicolumn{5}{c|}{\textbf{Assemble / Disassemble}} &
    \multicolumn{2}{c|}{\textbf{Articulated Object}} & \multirow{2}{*}{\textbf{Avg.}} \\
    \cmidrule(lr){2-6}\cmidrule(lr){7-11}\cmidrule(l){12-13}
     & \makecell{Apple  \\ $\to$   plate}
     & \makecell{ Bowl \\  $\to$  bowl}
     & \makecell{Two eggs \\ $\to$ box}
     & \makecell{Lift \\ basket}
     & \makecell{Left block \\ $\to$ right plate}
     & \makecell{Stack \\ ring blocks}
     & \makecell{Grab \\ square blocks}
     & \makecell{Place kettle \\ on base}
     & \makecell{Remove \\ pen cap}
     & \makecell{Separate \\ nested bowls}
     & \makecell{Open \\ cabinet door}
     & \makecell{Open \\ laptop} \\
    \midrule
    DP    & 60 & 65 & 30 & 10 & 25 & 35 & 15 & 45 & 30 & 10 & 65 & 20 & 34.2\\
    $\pi_0$   & 75 & 70 & 45 & 30 & 30 & 60 & 60 & 65 & 55 & 20 & 60 & 35 & 50.4\\
    GR00T N1 & 95 & \textbf{100} & 75 & 60 & 80 & \textbf{90} & \textbf{80} & 90 & 80 & 60 & 95 & 80 & 82.1\\
    \midrule
    Dexora & \textbf{100} & \textbf{100} & \textbf{85} & \textbf{80} & \textbf{90} & 85 & \textbf{80} & \textbf{95} & \textbf{90} & \textbf{80} & \textbf{100} & \textbf{90} & \textbf{89.6}\\
    \bottomrule
  \end{tabular}
  }
 \vspace{-10pt}
  \label{tab:basic_tasks}
\end{table*}

\begin{figure}
    \centering
    \includegraphics[width=1.0\linewidth]{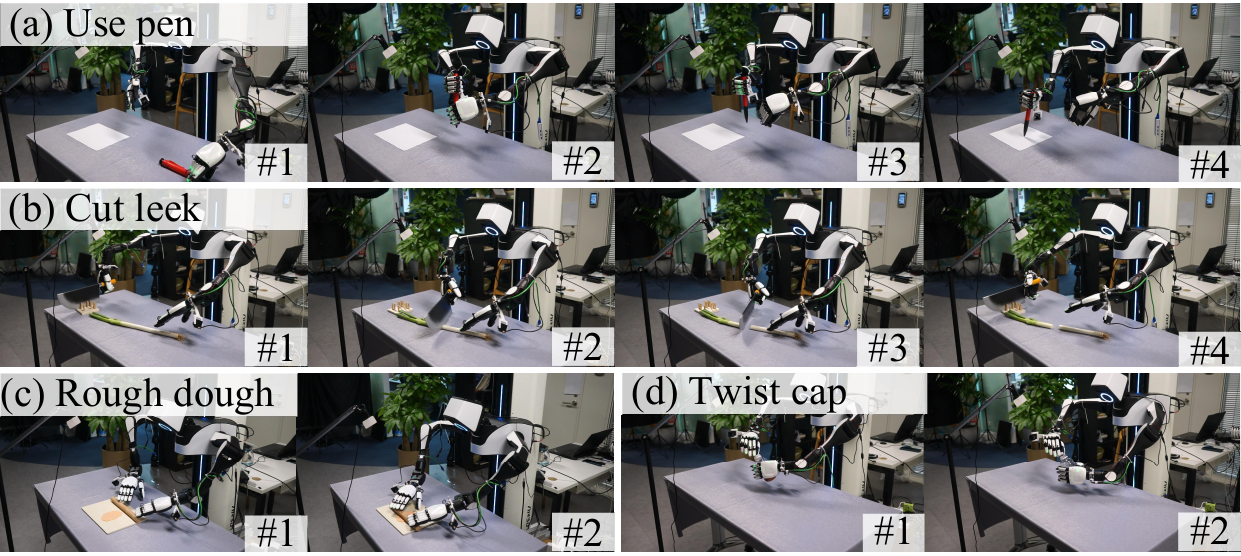}
    \caption{\textbf{Dexterous manipulation sequences}. (a) \textbf{Use Pen}: The left hand picks up the pen (\#1), hands it to the right hand (\#2); the right thumb depresses the tip (\#3) and writes on paper (\#4). (b) \textbf{Cut Leek}: The right hand grasps the knife (\#1), the left hand stabilizes the leek (\#2); the right hand slices (\#3) and returns the knife to the table (\#4).
(c) \textbf{Rough Dough}: Both hands press the rolling pin simultaneously (\#1) and push forward to flatten the dough (\#2).
(d) \textbf{Twist Cap}: The left hand holds the bottle while the right thumb–index grip twists the cap (\#1) and removes it (\#2).}
\vspace{-10pt}
    \label{fig:dex_manipulation}
\end{figure}

\textbf{Dexterous Manipulation Tasks Evaluation.} Pure pick-and-place does not exploit high-DoF hands; grippers can also do that. The value of hands emerges on dexterous skills that require in-hand tool use and coordinated bimanual manipulation (Fig.~\ref{fig:teaser}(a)). We therefore benchmark 6 tasks (Fig.~\ref{fig:dex_manipulation}). All models are trained/fine-tuned on 100 demonstrations. Tab.~\ref{tab:dex_tasks} shows that \emph{Dexora} gains the best average performance ($66.7\%$ vs.\ $51.7\%$ for GR00T~N1, $26.7\%$ for $\pi_{0}$, and $6.7\%$ for DP). GR00T~N1 is the strongest baseline but uses a 6-DoF hand; it struggles on in-hand skills such as \emph{Use pen} and fails on \emph{Twist cap}, which require thumb-index synergies and lateral finger swing to generate a stable torsional wrench. \emph{Dexora}'s gains arise from its 12-DoF hands and bimanual training corpus, enabling reliable in-hand and dual-arm coordination. We find that cap twisting exhibits the lowest success rate. The task requires generating a stable torsional wrench to overcome cap breakaway torque while preventing slip, which couples precise normal-force regulation, fingertip friction, and fine in-hand alignment. In our current setup, the absence of tactile feedback and relatively low-friction rigid fingertip pads leads to slip.

\begin{table}[t]
  \centering
  \small
   \caption{\textbf{Dexterous manipulation} tasks evaluation.}
  \setlength{\tabcolsep}{4.5pt}
  \renewcommand{\arraystretch}{0.95}
   \resizebox{\linewidth}{!}{
  \begin{tabular}{l|cccccc}
    \toprule
    Method & Use pen & Fetch book & Cut leek & Place plates & Rough dough& Twist cap \\
    \midrule
    DP        & 5 & 10 & 10 & 0 & 15  & 0\\
    $\pi_0$   & 20 & 45 & 60 & 20 & 15  & 0\\
    GR00T N1  & 45 & 60 & \textbf{85} & 60 & 60  & 0\\
    \midrule
    Dexora    & \textbf{65} & \textbf{80} & 80 & \textbf{70} & \textbf{80}  & \textbf{25}\\
    \bottomrule
  \end{tabular}
  }

  \label{tab:dex_tasks}
\end{table}

\subsection{Generalization} 
\label{sec: generalization}

\begin{figure}
    \centering
    \includegraphics[width=1.0\linewidth]{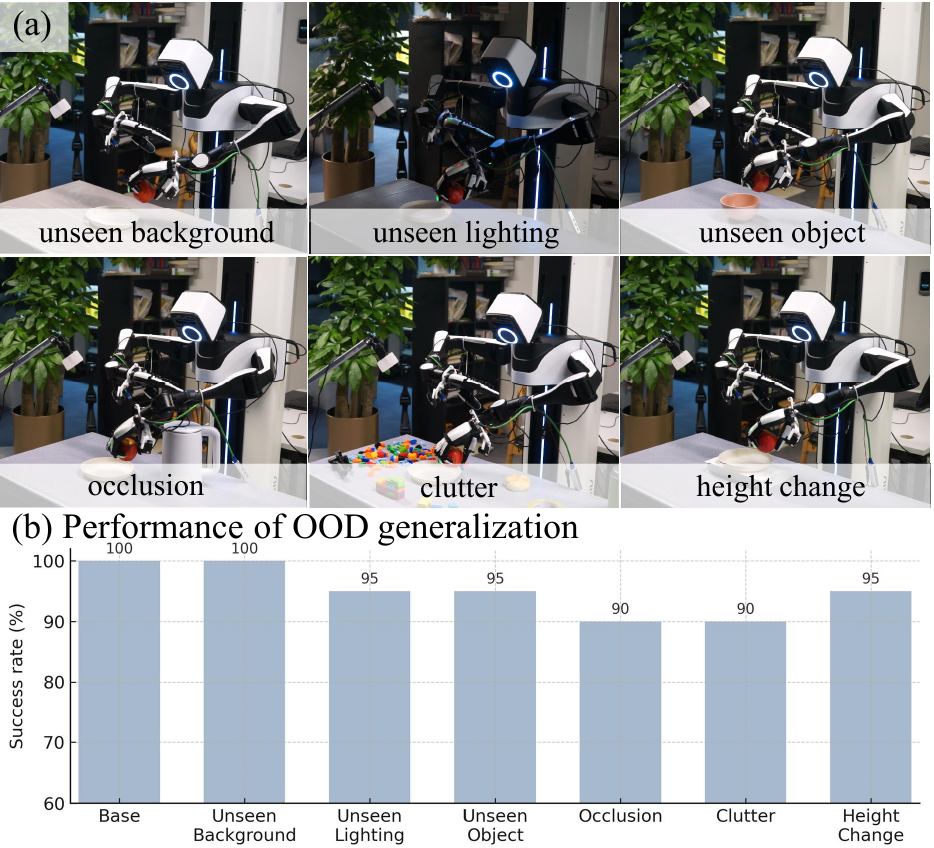}
    \caption{Generalization of six Out-of-Distribution (OOD) conditions. We report success rate (\%) over 20 rollouts.}
    \label{fig:OOD}
    \vspace{-0.5cm}
\end{figure}

\textbf{Out-of-Distribution Generalization.}
We test OOD robustness on the “Pick apple to the plate” task across six conditions: unseen background, unseen lighting, unseen object, occlusion, clutter, and height change, and we report the success rate (Fig.~\ref{fig:OOD}). \emph{Dexora} maintains high performance across all variants, showing excellent OOD generalization.

\begin{figure}
    \centering
    \includegraphics[width=1.0\linewidth]{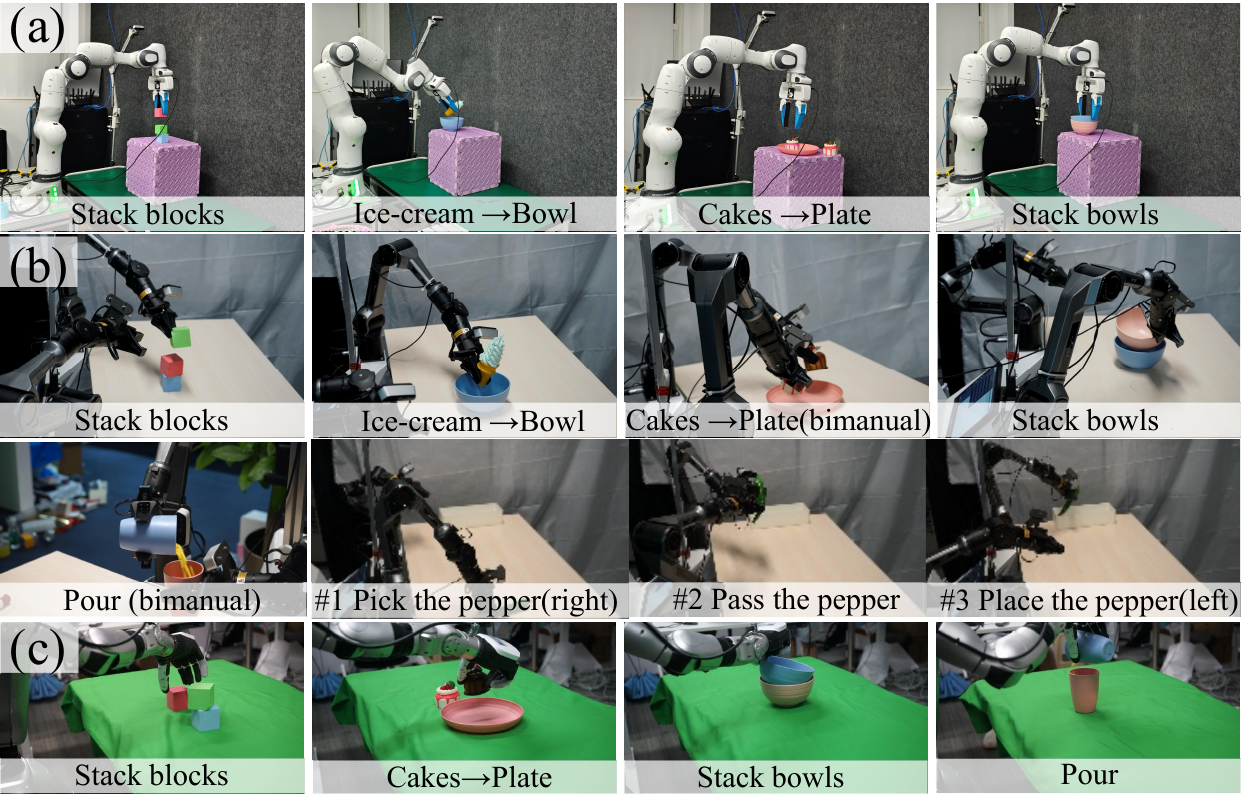}
    \caption{\textbf{Cross-embodiment generalization}. The \emph{Dexora} policy transfers to (a) single-arm gripper, (b) dual-arm grippers, and (c) single-arm single-hand, completing representative tasks like a three-step pepper handover. \vspace{2em}}
    \label{fig:cross-embodiment-gen}
    \vspace{-15pt}
\end{figure}

\textbf{Cross-Embodiment Generalization.}
Our premise is that a dual-arm, dual-hand high-DoF policy contains lower-DoF embodiments as subspaces: projecting a 36-D joint action down to simpler robots is dimension reduction, not synthesis—far easier than “lifting” a gripper policy to dexterous hands. We therefore test three representative embodiment configurations: \textbf{EC-1: single-arm gripper} - Franka Emika Panda (6-DoF + 1-DoF gripper); \textbf{EC-2: dual-arm grippers} - Cobot Magic ALOHA (2 × (6-DoF arm + 1-DoF gripper)); \textbf{EC-3: single-arm single-hand} - Unitree G1 7-DoF arm + Inspire Hand 6-DoF.
For adaptation, we pad unused action dimensions to keep tensor shapes fixed; for observations, we mask the absent camera. Each task is fine-tuned with 100 demonstrations, and all other settings are identical. On the evaluated tasks including single- and dual-arm setups (Fig.~\ref{fig:cross-embodiment-gen}), grasping tasks transfer readily across embodiments, whereas dexterity-demanding tasks show the largest gaps (Tab.~\ref{tab:dex_tasks}). This supports our hypothesis that high→low mapping is better posed than the inverse; compressing a 12-DoF hand policy to a 1-DoF gripper is simpler than lifting a gripper policy to dexterous hands.

\subsection{Ablation Study}
\label{sec: ablation study}
\begin{figure}
    \centering
    \includegraphics[width=0.95\linewidth]{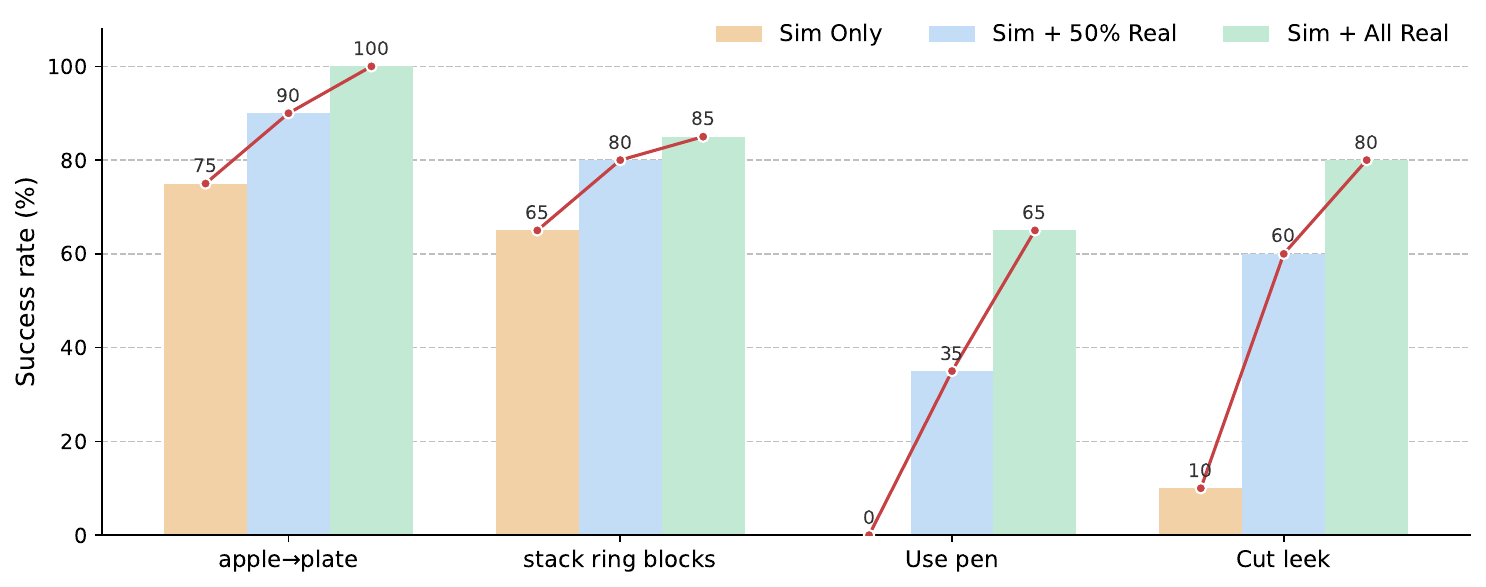}
    \caption{\textbf{Effect of training data composition.} Success rate for four tasks under three training regimes: Sim Only, Sim + 50\% Real, Sim + All Real.}
    \vspace{-20pt}
    \label{fig:scaling}
\end{figure}

\textbf{Effectiveness of Training Data Composition.} We compare three post-training regimes: Sim Only, Sim + 50\% Real (100 tasks), and Sim + All Real (200 tasks). Four tasks are evaluated, two basic (Apple→plate, Stack ring blocks) and two dexterous (Use pen, Cut leek). Success rises steadily with more real data; dexterous tasks improve from 0→35→65 and 10→60→85 (Fig.~\ref{fig:scaling}). These results show that simulation is effective for bootstrapping basic skills, while real, more complex data plays a crucial role in developing dexterous capabilities.

\begin{figure}
    \centering
    \includegraphics[width=1.0\linewidth]{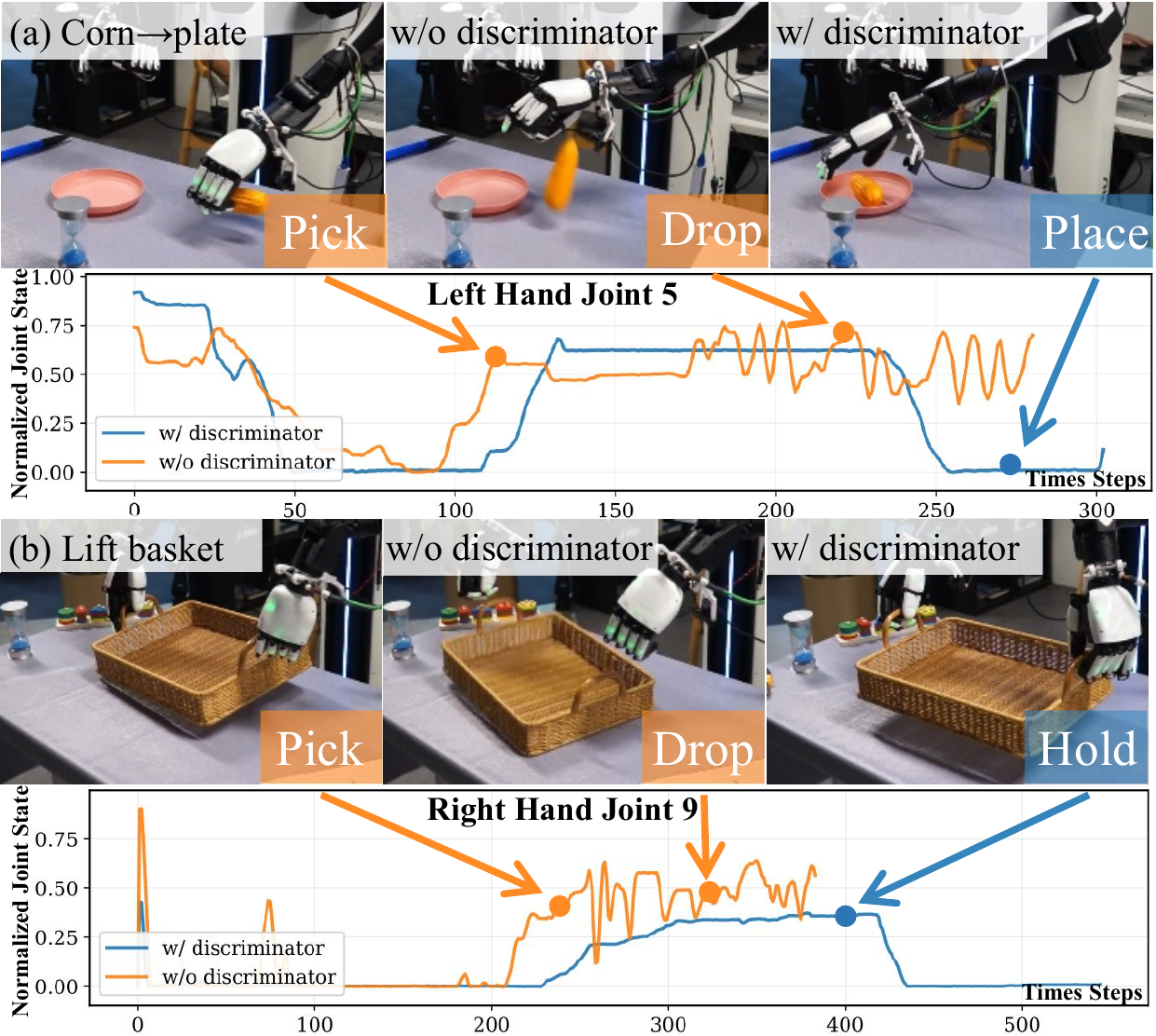}
    \caption{\textbf{Effect of the data-quality discriminator}. (a) \textbf{Corn $\rightarrow$ plate}: with the discriminator, joint trajectories are smooth and the placement succeeds; without it, high-frequency oscillations in left-hand joint 5 cause the corn to drop. (b) \textbf{Lift basket (bimanual)}: with the discriminator, the basket is lifted; without it, jitter in right-hand joint 9 tilts the basket and it slips.\vspace{2em}}
    \vspace{-15pt}
    \label{fig:discriminator}
\end{figure}

\vspace{10pt}

\begin{table}[t]
  \centering
  \small
  \caption{\textbf{Effect of the discriminator model}. We report \textbf{S.R.} (success rate \%) and smoothness metrics—mean normalized joint \textbf{Acceleration} and \textbf{Jerk}, averaged over 20 episodes. }
  \renewcommand{\arraystretch}{1.0}
  \resizebox{\linewidth}{!}{%
  \begin{tabular}{l|ccc|ccc}
    \toprule
    \multirow{2}{*}{\textbf{Method}} &
    \multicolumn{3}{c|}{\textbf{Corn $\to$ plate}} &
    \multicolumn{3}{c}{\textbf{Lift basket}} \\
    \cmidrule(lr){2-4}\cmidrule(lr){5-7}
     & S.R. & Acc. $\downarrow$ & Jerk $\downarrow$
     & S.R. & Acc. $\downarrow$ & Jerk $\downarrow$ \\
    \midrule
    w/o discriminator & 85 & 0.034 & 0.043 & 55 & 0.041 & 0.052 \\
     \rowcolor{gray!20}
    w/  discriminator & 95 & 0.020 & 0.032 & 80 & 0.023 & 0.036 \\
    \bottomrule
  \end{tabular}%
  }
  \label{Tab:ablation_discriminator}
  \vspace{-10pt}
\end{table}

\textbf{Effectiveness of Discriminator model.}
We compare vanilla post-training of the Diffusion Transformer with quality-aware post-training that uses a learned discriminator to score and weight demonstrations. Tab.~\ref{Tab:ablation_discriminator} quantifies the gains: the discriminator improves success rate and reduces acceleration and jerk at inference. In both a single-hand and a bimanual task, the quality-aware model executes smoother, more coherent motions. The time-series traces show lower variance and fewer reversals (Fig.~\ref{fig:discriminator}). Overall, the discriminator helps the policy learn from mixed-quality demonstrations by emphasizing high-quality segments and down-weighting suboptimal ones, enabling better strategies from imperfect data.

\section{CONCLUSION}
We present \emph{Dexora}, the first open-source VLA system that natively controls dual-arm, dual-hand, 36-DoF robots. A hybrid teleoperation pipeline drives both hardware and a MuJoCo twin to build an embodiment-matched corpus, and a data-quality discriminator guides post-training so the policy learns most from high-quality demonstrations. \emph{Dexora} outperforms strong baselines on basic and dexterous tasks, is robust to OOD shifts, and transfers across embodiments with lightweight action projectors—evidence that training in a rich, high-DoF action space provides a well-posed path to lower-DoF controllers. Ablations show that simulation bootstraps basic competence, while real data and the discriminator are key for dexterity and smooth control.

Looking forward, we see two promising directions: (i) contact-aware control via tactile sensing to close the loop on tasks like cap twisting; (ii) long-horizon reasoning and hierarchical VLA planning that combines memory, subgoal decomposition, and language-guided tool use. We hope the released models, data, and code catalyze research toward broadly capable, dexterous robot assistants.

{
\small
\bibliographystyle{ieeetr}
\balance
\bibliography{main}

@article{zhao2023learningfinegrained,
  title={Learning Fine-Grained Bimanual Manipulation with Low-Cost Hardware},
  author={Zhao, Tony Z. and Kumar, Vikash and Levine, Sergey and Finn, Chelsea},
  journal={RSS},
  year={2023}
}

@article{aloha22024enhanced,
  title={ALOHA 2: An Enhanced Low-Cost Hardware for Bimanual Teleoperation},
  author={ALOHA 2 Team},
  journal={arXiv preprint arXiv:2405.02292},
  year={2024}
}

@article{iyer2024openteach,
  title={OPEN TEACH: A Versatile Teleoperation System for Robotic Manipulation},
  author={Iyer, Aadhithya and Peng, Zhuoran and Dai, Yinlong and Guzey, Irmak and Haldar, Siddhant and Chintala, Soumith and Pinto, Lerrel},
  journal={CoRL},
  year={2024}
}

@article{ding2024bunnyvisionpro,
  title={Bunny-VisionPro: Real-Time Bimanual Dexterous Teleoperation for Imitation Learning},
  author={Ding, Runyu and Qin, Yuzhe and Zhu, Jiyue and Jia, Chengzhe and Yang, Shiqi and Yang, Ruihan and Qi, Xiaojuan and Wang, Xiaolong},
  journal={arXiv preprint arXiv:2407.03162},
  year={2024}
}

@article{qin2023anyteleop,
  title={AnyTeleop: A General Vision-Based Dexterous Robot Arm-Hand Teleoperation System},
  author={Qin, Yuzhe and Yang, Wei and Huang, Binghao and Van Wyk, Karl and Su, Hao and Wang, Xiaolong and Chao, Yu-Wei and Fox, Dieter},
  journal={RSS},
  year={2023}
}

@article{li2018teachnet,
  title={Vision-based Teleoperation of Shadow Dexterous Hand using End-to-End Deep Neural Network},
  author={Li, Shuang and Ma, Xiaojian and Liang, Hongzhuo and G{\"o}rner, Michael and Ruppel, Philipp and Fang, Bing and Sun, Fuchun and Zhang, Jianwei},
  journal={ICRA},
  year={2019}
}

@article{fang2023airexo,
  title={AirExo: Low-Cost Exoskeletons for Learning Whole-Arm Manipulation in the Wild},
  author={Fang, Hongjie and Fang, Hao-Shu and Wang, Yiming and Ren, Jieji and Chen, Jingjing and Zhang, Ruo and Wang, Weiming and Lu, Cewu},
  journal={ICRA},
  year={2024}
}

@article{xu2025dexumi,
  title={DexUMI: Using Human Hand as the Universal Manipulation Interface for Dexterous Manipulation},
  author={Xu, Mengda and Zhang, Han and Hou, Yifan and Xu, Zhenjia and Fan, Linxi and Veloso, Manuela and Song, Shuran},
  journal={CoRL},
  year={2025}
}

@article{imdieke2025sparkremote,
  title={SPARK-Remote: A Cost-Effective System for Remote Bimanual Robot Teleoperation},
  author={Imdieke, Adam and Desingh, Karthik},
  journal={arXiv preprint arXiv:2504.05488},
  year={2025}
}

@article{wu2023gello,
  title={GELLO: A General, Low-Cost, and Intuitive Teleoperation Framework for Robot Manipulators},
  author={Wu, Philipp and Shentu, Yide and Yi, Zhongke and Lin, Xingyu and Abbeel, Pieter},
  journal={IROS},
  year={2024}
}

@article{li2025demonstrationmodality,
  title={How to Train Your Robots? The Impact of Demonstration Modality on Imitation Learning},
  author={Li, Haozhuo and Cui, Yuchen and Sadigh, Dorsa},
  journal={arXiv preprint arXiv:2503.07017},
  year={2025}
}

@article{pan2024vladiffswitch,
  title={Vision-Language-Action Model and Diffusion Policy Switching Enables Dexterous Control of an Anthropomorphic Hand},
  author={Pan, Cheng and Junge, Kai and Hughes, Josie},
  journal={arXiv preprint arXiv:2410.14022},
  year={2024}
}

@article{zhang2025robustdexgrasp,
  title={RobustDexGrasp: Robust Dexterous Grasping of General Objects},
  author={Zhang, Hui and Wu, Zijian and Huang, Linyi and Christen, Sammy and Song, Jie},
  journal={arXiv preprint arXiv:2504.05287},
  year={2025}
}

@inproceedings{xu2023unidexgrasp,
  title={Unidexgrasp: Universal robotic dexterous grasping via learning diverse proposal generation and goal-conditioned policy},
  author={Xu, Yinzhen and Wan, Weikang and Zhang, Jialiang and Liu, Haoran and Shan, Zikang and Shen, Hao and Wang, Ruicheng and Geng, Haoran and Weng, Yijia and Chen, Jiayi and others},
  booktitle={CVPR},
  pages={4737--4746},
  year={2023}
}

@article{yang2025egovla,
  title={EgoVLA: Learning Vision-Language-Action Models from Egocentric Human Videos},
  author={Yang, Ruihan and Yu, Qinxi and Wu, Yecheng and Yan, Rui and Li, Borui and Cheng, An-Chieh and Zou, Xueyan and Fang, Yunhao and Yin, Hongxu and Liu, Sifei and others},
  journal={arXiv:2507.12440},
  year={2025}
}

@inproceedings{li2025maniptrans,
  title={Maniptrans: Efficient dexterous bimanual manipulation transfer via residual learning},
  author={Li, Kailin and Li, Puhao and Liu, Tengyu and Li, Yuyang and Huang, Siyuan},
  booktitle={CVPR},
  pages={6991--7003},
  year={2025}
}

@article{ye2025dex1b,
  title={Dex1B: Learning with 1B Demonstrations for Dexterous Manipulation},
  author={Ye, Jianglong and Wang, Keyi and Yuan, Chengjing and Yang, Ruihan and Li, Yiquan and Zhu, Jiyue and Qin, Yuzhe and Zou, Xueyan and Wang, Xiaolong},
  journal={arXiv preprint arXiv:2506.17198},
  year={2025}
}

@inproceedings{2025dexgraspanything,
  title={Dexgrasp anything: Towards universal robotic dexterous grasping with physics awareness},
  author={Zhong, Yiming and Jiang, Qi and Yu, Jingyi and Ma, Yuexin},
  booktitle={CVPR},
  pages={22584--22594},
  year={2025}
}

@article{jiang2024dexmimicgen,
  title={Dexmimicgen: Automated data generation for bimanual dexterous manipulation via imitation learning},
  author={Jiang, Zhenyu and Xie, Yuqi and Lin, Kevin and Xu, Zhenjia and Wan, Weikang and Mandlekar, Ajay and Fan, Linxi and Zhu, Yuke},
  journal={arXiv preprint arXiv:2410.24185},
  year={2024}
}

@article{chen2025dexonomy,
  title={Dexonomy: Synthesizing All Dexterous Grasp Types in a Grasp Taxonomy},
  author={Chen, Jiayi and Ke, Yubin and Peng, Lin and Wang, He},
  journal={arXiv preprint arXiv:2504.18829},
  year={2025}
}

@inproceedings{zhang2024dexgraspnet,
  title={Dexgraspnet 2.0: Learning generative dexterous grasping in large-scale synthetic cluttered scenes},
  author={Zhang, Jialiang and Liu, Haoran and Li, Danshi and Yu, XinQiang and Geng, Haoran and Ding, Yufei and Chen, Jiayi and Wang, He},
  booktitle={8th CoRL},
  year={2024}
}

@inproceedings{li2024semgrasp,
  title={Semgrasp: Semantic grasp generation via language aligned discretization},
  author={Li, Kailin and Wang, Jingbo and Yang, Lixin and Lu, Cewu and Dai, Bo},
  booktitle={ECCV},
  year={2024},
}

@inproceedings{ye2024g,
  title={G-hop: Generative hand-object prior for interaction reconstruction and grasp synthesis},
  author={Ye, Yufei and Gupta, Abhinav and Kitani, Kris and Tulsiani, Shubham},
  booktitle={CVPR},
  pages={1911--1920},
  year={2024}
}

@article{liu2024realdex,
  title={Realdex: Towards human-like grasping for robotic dexterous hand},
  author={Liu, Yumeng and Yang, Yaxun and Wang, Youzhuo and Wu, Xiaofei and Wang, Jiamin and Yao, Yichen and Schwertfeger, S{\"o}ren and Yang, Sibei and Wang, Wenping and Yu, Jingyi and others},
  journal={ arXiv:2402.13853},
  year={2024}
}

@article{2025dexgraspvla,
  title={Dexgraspvla: A vision-language-action framework towards general dexterous grasping},
  author={Zhong, Yifan and Huang, Xuchuan and Li, Ruochong and Zhang, Ceyao and Liang, Yitao and Yang, Yaodong and Chen, Yuanpei},
  journal={arXiv preprint arXiv:2502.20900},
  year={2025}
}

@article{bjorck2025gr00t,
  title={Gr00t n1: An open foundation model for generalist humanoid robots},
  author={Bjorck, Johan and Casta{\~n}eda, Fernando and Cherniadev, Nikita and Da, Xingye and Ding, Runyu and Fan, Linxi and Fang, Yu and Fox, Dieter and Hu, Fengyuan and Huang, Spencer and others},
  journal={arXiv preprint arXiv:2503.14734},
  year={2025}
}

@article{liu2024rdt,
  title={Rdt-1b: a diffusion foundation model for bimanual manipulation},
  author={Liu, Songming and Wu, Lingxuan and Li, Bangguo and Tan, Hengkai and Chen, Huayu and Wang, Zhengyi and Xu, Ke and Su, Hang and Zhu, Jun},
  journal={arXiv preprint arXiv:2410.07864},
  year={2024}
}

@article{kim2024openvla,
  title={Openvla: An open-source vision-language-action model},
  author={Kim, Moo Jin and Pertsch, Karl and Karamcheti, Siddharth and Xiao, Ted and Balakrishna, Ashwin and Nair, Suraj and Rafailov, Rafael and Foster, Ethan and Lam, Grace and Sanketi, Pannag and others},
  journal={arXiv preprint arXiv:2406.09246},
  year={2024}
}

@article{luo2025beingH0,
  title={Being-H0: Vision-Language-Action Pretraining from Large-Scale Human Videos},
  author={Luo, Hao and Feng, Yicheng and Zhang, Wanpeng and Zheng, Sipeng and Wang, Ye and Yuan, Haoqi and Liu, Jiazheng and Xu, Chaoyi and Jin, Qin and Lu, Zongqing},
  journal={arXiv preprint arXiv:2507.15597},
  year={2025}
}

@article{black2024pi_0,
  title={$\pi_0 $: A Vision-Language-Action Flow Model for General Robot Control},
  author={Black, Kevin and Brown, Noah and Driess, Danny and Esmail, Adnan and Equi, Michael and Finn, Chelsea and Fusai, Niccolo and Groom, Lachy and Hausman, Karol and Ichter, Brian and others},
  journal={arXiv preprint arXiv:2410.24164},
  year={2024}
}

@article{intelligence2025pi_05,
  title={$\pi_ {0.5}$: a Vision-Language-Action Model with Open-World Generalization},
  author={Intelligence, Physical and Black, Kevin and Brown, Noah and Darpinian, James and Dhabalia, Karan and Driess, Danny and Esmail, Adnan and Equi, Michael and Finn, Chelsea and Fusai, Niccolo and others},
  journal={arXiv preprint arXiv:2504.16054},
  year={2025}
}

@article{gr3,
  title={GR-3 Technical Report},
  author={Cheang, Chilam and Chen, Sijin and Cui, Zhongren and Hu, Yingdong and Huang, Liqun and Kong, Tao and Li, Hang and Li, Yifeng and Liu, Yuxiao and Ma, Xiao and others},
  journal={arXiv preprint arXiv:2507.15493},
  year={2025}
}

@article{bu2025agibot,
  title={Agibot world colosseo: A large-scale manipulation platform for scalable and intelligent embodied systems},
  author={Bu, Qingwen and Cai, Jisong and Chen, Li and Cui, Xiuqi and Ding, Yan and Feng, Siyuan and Gao, Shenyuan and He, Xindong and Hu, Xuan and Huang, Xu and others},
  journal={arXiv preprint arXiv:2503.06669},
  year={2025}
}

@inproceedings{zitkovich2023rt2,
  title={Rt-2: Vision-language-action models transfer web knowledge to robotic control},
  author={Zitkovich, Brianna and Yu, Tianhe and Xu, Sichun and Xu, Peng and Xiao, Ted and Xia, Fei and Wu, Jialin and Wohlhart, Paul and Welker, Stefan and Wahid, Ayzaan and others},
  booktitle={CoRL},
  year={2023},
  organization={PMLR}
}

@article{jang2025dreamgen,
  title={DreamGen: Unlocking Generalization in Robot Learning through Neural Trajectories},
  author={Jang, Joel and Ye, Seonghyeon and Lin, Zongyu and Xiang, Jiannan and Bjorck, Johan and Fang, Yu and Hu, Fengyuan and Huang, Spencer and Kundalia, Kaushil and Lin, Yen-Chen and others},
  journal={},
  pages={arXiv--2505},
  year={2025}
}

@article{2025graspvla,
  title={Graspvla: a grasping foundation model pre-trained on billion-scale synthetic action data},
  author={Deng, Shengliang and Yan, Mi and Wei, Songlin and Ma, Haixin and Yang, Yuxin and Chen, Jiayi and Zhang, Zhiqi and Yang, Taoyu and Zhang, Xuheng and Cui, Heming and others},
  journal={arXiv preprint arXiv:2505.03233},
  year={2025}
}

@inproceedings{xu2022discriminator,
  title={Discriminator-weighted offline imitation learning from suboptimal demonstrations},
  author={Xu, Haoran and Zhan, Xianyuan and Yin, Honglei and Qin, Huiling},
  booktitle={International Conference on Machine Learning},
  pages={24725--24742},
  year={2022},
  organization={PMLR}
}

@article{chi2023diffusionpolicy,
  title={Diffusion policy: Visuomotor policy learning via action diffusion},
  author={Chi, Cheng and Xu, Zhenjia and Feng, Siyuan and Cousineau, Eric and Du, Yilun and Burchfiel, Benjamin and Tedrake, Russ and Song, Shuran},
  journal={IJRR},
  year={2023},
}

@article{T5,
  title={Exploring the limits of transfer learning with a unified text-to-text transformer},
  author={Raffel, Colin and Shazeer, Noam and Roberts, Adam and Lee, Katherine and Narang, Sharan and Matena, Michael and Zhou, Yanqi and Li, Wei and Liu, Peter J},
  journal={JMLR},
  volume={21},
  pages={1--67},
  year={2020}
}

@inproceedings{siglip,
  title={Sigmoid loss for language image pre-training},
  author={Zhai, Xiaohua and Mustafa, Basil and Kolesnikov, Alexander and Beyer, Lucas},
  booktitle={Proceedings of the IEEE/CVF international conference on computer vision},
  pages={11975--11986},
  year={2023}
}

@article{deitke2023objaverse,
  title={Objaverse-xl: A universe of 10m+ 3d objects},
  author={Deitke, Matt and Liu, Ruoshi and Wallingford, Matthew and Ngo, Huong and Michel, Oscar and Kusupati, Aditya and Fan, Alan and Laforte, Christian and Voleti, Vikram and Gadre, Samir Yitzhak and others},
  journal={NeurIPS},
  volume={36},
  year={2023}
}

@article{bai2025qwen2,
  title={Qwen2. 5-vl technical report},
  author={Bai, Shuai and Chen, Keqin and Liu, Xuejing and Wang, Jialin and Ge, Wenbin and Song, Sibo and Dang, Kai and Wang, Peng and Wang, Shijie and Tang, Jun and others},
  journal={arXiv preprint arXiv:2502.13923},
  year={2025}
}

@article{zhang2025tavla,
  title={Ta-vla: Elucidating the design space of torque-aware vision-language-action models},
  author={Zhang, Zongzheng and Xu, Haobo and Yang, Zhuo and Yue, Chenghao and Lin, Zehao and Gao, Huan-ang and Wang, Ziwei and Zhao, Hao},
  journal={arXiv preprint arXiv:2509.07962},
  year={2025}
}

@article{zhang2025robochemist,
  title={RoboChemist: Long-Horizon and Safety-Compliant Robotic Chemical Experimentation},
  author={Zhang, Zongzheng and Yue, Chenghao and Xu, Haobo and Liao, Minwen and Qi, Xianglin and Gao, Huan-ang and Wang, Ziwei and Zhao, Hao},
  journal={arXiv preprint arXiv:2509.08820},
  year={2025}
}
}
\balance

\end{document}